# Generalized-TODIM Method for Multi-criteria Decision Making with Basic Uncertain Information and Its Application

Zhiyuan Zhou[3], Kai Xuan[5], Zhifu Tao[1, 2], Ligang Zhou[3, 4]

[1] School of Economics, Anhui University, Hefei, Anhui, 230601, China

[2] Research Center for Finance and Statistics, Anhui University, Hefei, Anhui, 230601, China

[3] School of Mathematical Sciences, Anhui University, Hefei, Anhui, 230601, China

[4] Research Center for Applied Mathematics, Anhui University, Hefei, Anhui, 230601, China

[5] College of Letters and Science, University of California, Los Angeles, California, 90024, USA

**Abstract**: Due to the fact that basic uncertain information provides a simple form for decision information with certainty degree, it has been developed to reflect the quality of observed or subjective assessments. In order to study the algebra structure and preference relation of basic uncertain information, we develop some algebra operations for basic uncertain information. The order relation of such type of information has also been considered. Finally, to apply the developed algebra operations and order relations, a generalized TODIM method for multi-attribute decision making with basic uncertain information is given. The numerical example shows that the developed decision procedure is valid.

### 1.Introduction

The methods for solving multi-criteria decision-making problems had experienced drastic development because of the wild extension of its implementation environment for few decades. In contemporary times, this kind of methodology has been considered the indispensable tool in resolution in many decision-making applications where various and different criteria should be considered in the real world. Up to now, an amount of multi-criteria decision-making approaches have been proposed. For instance, the MULTIMOORA (Hafezalkotob, 2019), the MABAC method (Jia, 2019), the TODIM method (Zhang et al., 2019), the VIKOR method (Wang, 2018; Manna, 2020), and the PROMETHEE method (Liu 2019b). The TODIM (an acronym in Portuguese of interactive and multi-criteria decision making) method, proposed by Gomes and Lima, is an effective method based on the prospect theory for capturing DM's psychological behaviors. The TODIM method has received increased attention from the scientific community (Passos and Gomes (2014), Passos, Teixeira, Garcia, Cardoso, and Gomes (2014), and Lee and Shih (2016)), and it has been applied to different multi-criteria problems. For example, the TODIM method has been used to evaluate residential properties available for rent in the city of Volta Redonda, Brazil (Gomes & Rangel, 2009), to select the best option for the location of the natural gas reserves discovered in the Santos basin, Brazil (Gomes, Rangel & Maranhão, 2009), and to rank chemical industries and Thermal Power Station Units (Soni, Christian, & Jariwala, 2016).

Recently, some examples show that this method is vulnerable to two paradoxes concerning the weights of the model. in order to avoid the occurrence of these two paradoxes, Llamazares carry out an analysis of the TODIM method for introducing two properties called weight consistency and weight monotonicity. Based on that, he proposes a generalization of the TODIM method and gives conditions under which the generalized TODIM method satisfies weight consistency and weight monotonicity.

However, the information from assessment or observations of these existing approaches usually has some invisible error, i.e., artificial error and systematical error. Therefore, the uncertainty insides this decision process needs to be considered.

Fuzzy set theory is an effective tool to deal with ambiguity and uncertainty in decision-making. Zadeh and Goguen proposed the fuzzy set and L-fuzzy set in 1965 and 1967, respectively. Since the introduction of fuzzy sets by Zadeh, fuzzy sets have been successfully used to solve ambiguities and uncertainties, which are many practical problems encountered in various fields (such as decision-making, pattern recognition, clustering, and fuzzy topology. Subsequently, multiple extensions of fuzzy sets, such as interval-valued fuzzy sets, intuitionistic fuzzy sets (IFS), 2-type fuzzy sets, n-type fuzzy sets. The intuitionistic fuzzy set proposed by Atanassov considers membership degree, non-membership degree, and hesitation degree simultaneously. It is more objective and practical to express the fuzzy uncertainty in decision-making than the fuzzy set that only considers membership degree. Torra and Narukawa recently proposed Hesitating Fuzzy Sets (HFS) to describe the situation where the membership of an element to a given set has several different values. This situation will occur in many practical cases. It allows an object to belong to a fuzzy set to be shown in the form of a set of multiple possible values, unlike other fuzzy sets that require experts to give an error range or a distribution of several possible values for the attribute value. Therefore, hesitant fuzzy sets can effectively describe the uncertainty in decision-making. Further, to better avoid the problem of information loss in the decision-making process, and based on the characteristics of hesitant fuzzy sets and the flexibility and practicality of interval number expression, Cai proposed the definition of interval-valued hesitant fuzzy sets (IVHFS). Simultaneously, some operations and properties of interval-valued hesitant fuzzy elements are given, and the relationship between interval-valued hesitant fuzzy sets and several other fuzzy sets is explained.

Recently, Z-numbers (Zadeh, 2011) concepts and basic uncertain information (Mesiar et al., 2017) develop the description of credibility for uncertain information to improve the quality of information. Unlike Z-numbers' notion, basic uncertain information (BUI) (Mesiar et al., 2017) provides an effective and straightforward way to describe assessments or observations with a certain degree but not the restriction on the probability measure of the first component. It can also be seen that the degree of certainty provides a measure of the quality of information.

Since the BUI had been introduced, many scholars have discussed its aggregation approaches and operational rules. Furthermore, theoretical and fundamental research, including basic set operations, similarities, and differences of each BUI and its applications, has not been fully established. Besides, how to cope with the uncertainty from actual group decision making, which is represented by decision information vertical and structures, is an interesting and open question when facing MCGDM problems under basic uncertain information environment. To solve multi-criteria group decision-making from the two aforementioned perspectives, the combination of generalized TODIM and basic uncertain information would be put forward and illustrated how it applied to the realistic circumstance.

The cartesian partial order if the Mesair has given BUI, but the property he provided has no enough to illustrate how to determine the total order between any two different BUI when elements inside have not strictly followed the rules he mentioned. Thus, to develop the concept of generalized TODIM under a basic uncertain environment, we utilize relations between interval-number and BUI-value and the possibility of possibility degree of interval-number to help obtain approximate total order for solving this problem. Furthermore, we would put forward a novel value function that satisfied the two consistency properties of generalized TODIM under the basic uncertain information environment. Based on that, a novel procedure of MCDM problems is obtained. Theoretically, some approximate algebraic structure of BUI would be proposed, and the benefits of the solution of problems of MCDM with BUI inputs would also be discussed and provided. While in practice, the Covid-19 outbreak has threatened the lives of thousands of people in different countries and slowed down many industries' economic livelihoods in China. Assessments on cities' biosafety in central China come to be a critical indicator to prevent the spreading of the epidemic. Historical data can be collected from CDC documents and paper-based files, but the conclusions might not be the same. As a result, the assessments would be provided in the form of BUI. Correspondingly, an assessment of cities' biosecurity in central China based on MCGDM with BUI would be presented.

to reach the aims mentioned above, the main motivations of this paper will be structured as follow:

Section 2 mainly lists the necessary notions and notations, including basic uncertain information and generalized TODIM approaches. In Section 3, the approximate algebraic structure of BUI is studied. The properties of the operations are also given. While in Section 4, the possibility degree of BUI is considered. Group decision algorithms via basic uncertain information generalized TODIM approach is shown in Section 5. in Section 6, a numerical example of evaluating the alarming biosafety system of cities in central China is shown to illustrate the feasibility and validity of the developed group decision procedure. At last, the conclusions and remarks on possible future work are summarized in section 7.

## 2. Relations between BUI and interval number

in order to calculate the possibility degree of BUI value, we roughly define some kinds of binary calculating principle which stretched from the relations between interval numbers and BUI value defined by masier(2017)

According to the Mesiar(2017), when the interval value $[a,b]$ has been fixed, the uncertainty degree of any BUI value can be quantified by its length. Therefore the relations between BUI value and interval number, which is any single BUI-value can be represented as closed interval value in different ways, can be derived. Also, he proposes that $[a,b]=[cx,cx+c-1]$ in order to create the proportional link between and $[a,b]$ which dominated by the datum $x$. formally, the idea mentioned above also respectively adopts circumstance when $x=0\ or\ x=1$. in more generalized perspective, he introduces mapping $\phi:J\to[0,1], \phi(<x;c>)=[cx,cx+1-c]$ which is a bijection from sets $J\setminus\{<x;0>\mid x\in[0,1]\}$ to $L([0,1])\setminus\{[0,1]\}$ (also can be seen as stretch as BUI axiom 3). And the interval number [a,b] also can be assigned back to BUI value

$<\dfrac{a}{1-b+a}; 1-b+a>$ (with the convention $0/0=1$) by utilizing inverse mapping $\phi^{-1}$.

By using mapping $\phi$ and $\phi^{-1}$, we can roughly define a novel calculating operator between two BUI values.

Proposition k: if $A=<x_1;c_1>$, $B=<x_2;c_2>$ are any two BUI, considering $C=<x_3;c_3>$. we define addition operator "$\oplus$" and let $C=A\oplus B$. we can derive that

$$C=<\dfrac{c_1x_1+c_2x_2}{|c_1+c_2-1|}; |c_1+c_2-1|>$$

Proof: assuming that $A_1, B_1$ are interval number transformed from A and B, which is

$A_1=\phi(A)$, $B_1=\phi(B)$. According to the definition k, we notice that the lower boundary of $A_1$, or $B_1$ can be shown as $c_ix_i+c_i-1$ $(i\in 1,2)$. by the same principle, we can easily obtain the upper boundary as $c_ix_i$ $(i\in 1,2)$.

hence, based on calculating the principle of interval number, we can derive that considering interval number $C_1$, which is transformed from a BUI value $C=<x_3;c_3>$,

$C_1=\phi(C) \Leftrightarrow C=\phi^{-1}(C_1)$.

let $C_1=A_1+B_1$, with respect to the definition k, $C_1$ can be represented as $C_1=[c_3x_3, c_3x_3+c_3-1]$. and we notice that

$$C = \phi^{-1}(C_1) = \phi^{-1}(A_1 + B_1) = \phi^{-1}(\phi^{-1}(A_1) + \phi^{-1}(B_1)) = A \oplus B,$$

therefore, we can obtain the system of equations as follow:

$$\begin{cases} c_1 x_1 + c_2 x_2 = c_3 x_3 \\ c_1 x_1 + c_2 x_2 + 2 - (c_1 + c_2) = c_3 x_3 - c_3 + 1 \end{cases}$$

by solving these equations, we can find that $c_3$ they can be represented as $c_1 + c_2 - 1$

and $x_3$ can be represented as $\dfrac{c_1 x_1 + c_2 x_2}{c_1 + c_2 - 1}$.

consequently, the C can be derived as

$$C = <\dfrac{c_1 x_1 + c_2 x_2}{|c_1 + c_2 - 1|}; |c_1 + c_2 - 1|>$$

remark k: the solutions of equation k1, k2 are only meaningful in the mathematical field, but in solving the MCDM problems, the elements of BUI value can't below zero. thus, we need to convert $x_3$ and $c_3$ into the absolute value.

following the same principle, we can define the subtraction operator "$\ominus$"

proposition k: let $A = <x_1; c_1>$, $B = <x_2; c_2>$ are any two BUI, considering $C = <x_3; c_3>$ and let C=A$\ominus$B, we can derive that

$$C = <\left|\dfrac{c_1 x_1 - c_2 x_2 + c_2 - 1}{c_1 + c_2 - 1}\right|; |c_1 + c_2 - 1|>$$

proof: following the same principle of the proof of proposition k, the lower boundary of

C can be represented as $c_1x_1 - c_2x_2 + c_2 - 1$ an upper bound can be shown as $c_1x_1 - c_2x_2 - c_1 + 1$. Thus, the equation system k can be held as:

$$\begin{cases} c_1x_1 - c_2x_2 + c_2 - 1 = c_3x_3 \\ c_1x_1 - c_2x_2 - c_1 + 1 = c_3x_3 - c_3 + 1 \end{cases}$$

By solving these equations, the numerical solution of $c_3$ and $x_3$ can be obtained. therefore, we can derive that

$$C = < \left| \frac{c_1x_1 - c_2x_2 + c_2 - 1}{c_1 + c_2 - 1} \right|; |c_1 + c_2 - 1| >$$

Remark k: as same as proposition k, when implementing in some specific MCDM problems, elements of C should be converted into absolute value.

## 3. outranking relations of basic uncertain information

In the next two closely connected subsections, we propose a novel concept named possibility degree of basic uncertain information. Moreover, the outranking relations of two different BUI are defined on the basis of possibility degree of interval-numbers.

### 3.1 possibility degree of BUI

definition k: considering any two BUI $A = <x_1; c_1>$, $B = <x_2; c_2>$ and which any elements insides $\{x_1, x_2, c_1, c_2\}$ is a real number. Therefore, some comparative ideas can be defined as follows:

1. if $x_1 = x_2, c_1 = c_2$, then we call A is absolutely equal to B and marked it as A= B

2. if $x_1 > x_2, c_1 > c_2$, then we call A is strictly larger than B and marked it as A>B

remark k: current cartesian partial order given by mesair has not satisfied in any circumstance and cannot distinguish the comparison between any two BUI-value. In light of that, without losing any generality, we consider using the transformation between interval-number and BUI to calculate the possibility degree of BUI-value.

**3.2 calculating the possibility degree of BUI**

definition: let $A =< x_1; c_1 >$, $B =< x_2; c_2 >$ be any two BUI-value, the algorithm for calculating the possibility degree of two BUI values A and B can be defined as follow:

Step1. Transforming the A, B into its interval-value counterpart $A_1$ $B_1$ by using the equation mentioned above:

$$A_1 = [c_1 x_1, c_1 x_1 - c_1 + 1] \qquad B_1 = [c_2 x_2, c_2 x_2 - c_2 + 1]$$

Step2. Calculating the possibility degree between A1 and B1 by utilizing the formula proposed by (Gao 2013; Xu 2001; Xu and Da 2003), which is:

$$P(A_1 \geq B_1) = \max\{1 - \max(\frac{b^+ - a^-}{(b^+ - b^-) + (a^+ - a^-)}, 0), 0\}$$

and we derive that

$$P(A_1 \geq B_1) = \max\{1 - \max(\frac{c_1 x_1 + c_2 x_2 - c_1 + 1}{2 - (c_1 + c_2)}, 0), 0\}$$

herein, the possibility degree of A1 and B1 can be considered as the approximately equivalent of that in BUI value A and B, which is:

$$P(A \geq B) = \max\{1 - \max(\frac{c_1 x_1 - c_2 x_2 - c_1 + 1}{2 - (c_1 + c_2)}, 0), 0\}$$

definition k(Gao,2013): considering A=[a-,a+] (a-a+), B=[b-,b+] and C=[c-,c+] be any three interval numbers, some properties of possibility degree of them can be defined as follow:

1. Normative: $0 \leq p(A \geq B) \leq 1$;

2. Complementary: $p(A \geq B) + p(b \geq a) = 1$;

3. Reflexivity: $p(A \geq B) = p(B \geq A) = 0.5$ if A=B;

4. Transitivity: if $p(A \geq B) \leq 0.5$ and $p(B \geq C) \leq 0.5$, then $p(A \geq C) \leq 0.5$.

inspired by the definition k, we propose the concept of possibility degree of BUI-value.

definition k: let $A = <x_1; c_1>$, $B = <x_2; c_2>$ be any two BUI value, some comparative relations can be derived as follow:

1. if $P(A \geq B) + P(B > A) = 1$, then we call A is indifferent with B and marked as $A \square B$

2. if $P(A \geq B) > 0.5$, then we call A or B is weakly smaller other one and marked as $A \succ B$

Theorem k: Let any two $A = <x_1; c_1>$, $B = <x_2; c_2>$ be any two BUI. Some properties of the possibility degree of BUI-value are satisfied as follows:

1. Normative: $0 \leq p(A \geq B) \leq 1$;

2. Complementary: $p(A \geq B) + p(B \geq A) = 1$;

3. Reflexivity: $p(A \geq B) = p(B \geq A) = 0.5$ if A= B.

1. Proof: the essential possibility degree of BUI value can be seen as counterpart one in interval value. Therefore, due to the defined k, the normative of p(A≥B) is equal to p(A1≥B1) (A1, B1 are interval numbers transformed from A and B), which is $0 \leq p(A \geq B) \leq 1$.

Property 1 has been proved.

2. Proof: there are three cases that exist.

for case 1, we assume that if the outcome of equation k is

$$P(A \geq B) = 1 - \frac{c_1 x_1 - c_2 x_2 - c_1 + 1}{2 - (c_1 + c_2)}$$

follow property 1; we can derive that

$$0 < 1 - \frac{c_1 x_1 - c_2 x_2 - c_1 + 1}{2 - (c_1 + c_2)} < 1$$

Which is equally as $c_2 x_2 - c_1 x_1 - c_2 + 1 > 0$. Also, we notice that $2 - (c_1 + c_2) > 0$. thus, we can find that $\frac{c_2 x_2 - c_1 x_1 - c_2 + 1}{2 - (c_1 + c_2)} > 0$ and by studying the equation k, we notice that $\frac{c_2 x_2 - c_1 x_1 - c_2 + 1}{2 - (c_1 + c_2)} < 1$.

consequently, the outcome of $P(B > A) = \max\{1 - \max(\frac{c_2 x_2 - c_1 x_1 - c_2 + 1}{2 - (c_1 + c_2)}, 0), 0\}$

is $1 - \frac{c_2 x_2 - c_1 x_1 - c_2 + 1}{2 - (c_1 + c_2)}$, which is complementary of $P(A \geq B)$.

thus,

$$P(A \geq B) + P(B > A) = 1 - \frac{c_2 x_2 - c_1 x_1 - c_2 + 1}{2 - (c_1 + c_2)} + 1 - \frac{c_1 x_1 - c_2 x_2 - c_1 + 1}{2 - (c_1 + c_2)} = 2 - \frac{2 - c_1 - c_2}{2 - (c_1 + c_2)} = 1$$

for case 2, we assume the outcome of equation k be 0; then it's easy to note that

$$\frac{c_1 x_1 - c_2 x_2 - c_1 + 1}{2 - (c_1 + c_2)} > 1 \Leftrightarrow c_2 x_2 - c_1 x_1 - c_2 + 1 < 0$$

therefore, as the complement, $P(B > A)$ it can be derived as

$$P(B > A) = \max\{1 - \max(\frac{c_2 x_2 - c_1 x_1 - c_2 + 1}{2 - (c_1 + c_2)}, 0), 0\} = 1$$

Hereinafter, $P(A \geq B) + P(B > A) = 0 + 1 = 1$.

For case 3, let the outcome of equation k be 1, suggesting that:

$$c_1 x_1 - c_2 x_2 - c_1 + 1 < 0 \Leftrightarrow c_2 x_2 - c_1 x_1 - c_2 + 1 > 2 - (c_2 + c_1) \Rightarrow \frac{c_2 x_2 - c_1 x_1 - c_2 + 1}{2 - (c_1 + c_2)} > 1$$

thus, we can note that $P(B > A) = 1 \Leftrightarrow P(A \geq B) + P(B > A) = 1 + 0 = 1$

the second property has been proved.

3. proof: if $P(A \geq B) = 0.5$, then

$$P(A \geq B) = \max\{1 - \max(\frac{c_1 x_1 - c_2 x_2 - c_1 + 1}{2 - (c_1 + c_2)}, 0), 0\} = 0.5 \Leftrightarrow c_1 x_1 - c_2 x_2 - c_1 + 1 = \frac{1}{2}(c_1 + c_2)$$

$$\Rightarrow c_2 x_2 - c_1 x_1 = \frac{1}{2}(c_1 - c_2)$$

thus,

$$P(B > A) = \max\{1 - \max(\frac{c_2 x_2 - c_1 x_1 - c_2 + 1}{2 - (c_1 + c_2)}, 0), 0\} = \frac{2 - (c_1 + c_2) + (c_1 x_1 - c_2 x_2) + c_2 - 1}{2 - (c_1 + c_2)}$$

$$= \frac{1 - (c_1 + c_2) + \frac{1}{2}(c_1 - c_2) + c_2}{2 - (c_1 + c_2)} = \frac{1 - \frac{1}{2}(c_1 + c_2)}{2 - (c_1 + c_2)} = 0.5$$

hence, $P(A \geq B) = P(B > A) \Leftrightarrow A = B$

the reflexivity property has been proved.

## 4. solving MCDM problems via BUI-GTODIM method

in this section, we apply the BUI-GTODIM method for solving the multicriteria decision making problems under an uncertain environment.

Assuming the A={a1,a2,…,an} be a finite set of alternatives and let C={c1,c2…,cn} be the set of criteria in the problem K. we use the n,m to note the indices of BUI value under the different criteria. The performance of different alternatives in accordance to every criterion is given by experts via BUI-value.

Supposing that multicriteria method A is a function from D into W, where W is the set of weak orders (complete and transitive binary relations) on A.

Let w={w1,w2,…,wn} be a normalized vector which associated criteria k and reflect the critical degree of that and elements of w are normalized that satisfied $\sum_{i=1}^{n} w_i = 1$.

Hereinafter, we can define the relative weight of criterion $c_k$ over another criterion $c_r$,

that is: $w_{kr} = \frac{w_k}{w_r}$.

The core of the classical TODIM method is the prospect value function(Fig.1.) which provides the dominance for every criterion in accordance to each alternative. as in the prospect theory in (Bonifacio Llamazares, 2017), the value function of generalized TODIM should satisfy two key properties: the weight consistency and monotonicity. Furthermore, notice that the input of performance $z_{ij}$ is BUI-value, suggesting that the value function should hold the definition that the BUI aggregation function has. Before we introduce the generalized BUI TODIM method, some simplification processes should be mentioned first.

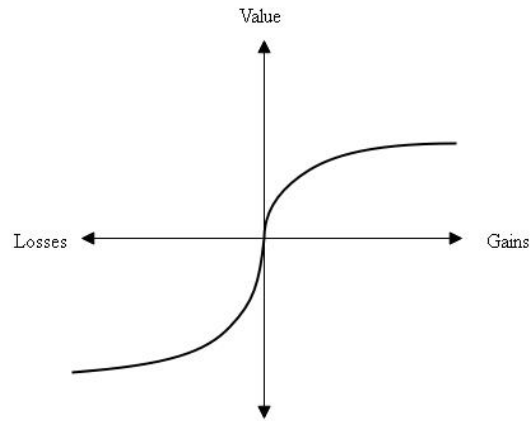

Fig. 1. prospect value function

Definition k: mark the $\varphi_k(A_i, A_j)$ as the value function which reflects the dominance degree of alternatives $A_j$ over $A_i$ with respect to criteria $c_k$.

according to the Gomes and Lima(2005), the form of classical value functions of TODIM can be written as

$$\varphi_k(A_i, A_j) = \begin{cases} \left( \dfrac{w_{kr}(z_{ik} - z_{jk})}{\sum_{l=1}^{m} w_{lr}} \right)^a & if \ z_{ik} \geq z_{jk} \\ -\lambda \left( \dfrac{\sum_{l=1}^{m} w_{lr}(z_{ik} - z_{jk})}{w_{kr}} \right)^a & if \ z_{jk} \geq z_{ik} \end{cases}$$

And the convexity or concavity of the of value function is determined by the estimable parameter $a$ and $b$, which satisfied $0<a,b<1$. $\lambda>1$ is called the loss aversion coefficient of function.

for promotingly simplifying the equation system k, Bonifacio Llamazares(2017) find that, for any $k \in M$

$$\frac{w_{kr}}{\sum_{l=1}^{m} w_{lr}} = \frac{\frac{w_k}{w_r}}{\sum_{l=1}^{m} \frac{w_l}{w_r}} = \frac{w_k}{\sum_{l=1}^{m} w_l} = w_k$$

so that:

$$\varphi_k(A_i, A_j) = \begin{cases} \left(w_k(z_{ik} - z_{jk})\right)^a & \text{if } z_{ik} \geq z_{jk} \\ -\lambda\left(\frac{(z_{ik} - z_{jk})}{w_k}\right)^a & \text{if } z_{jk} \geq z_{ik} \end{cases}$$

in a more generalized perspective, Bonifacio proposes that

$$\varphi_k(A_i, A_j) = \begin{cases} g_1(w_k)f_1(\chi_{jk}) & \text{if } z_{ik} \geq z_{jk} \\ -g_2(w_k)f_2(\chi_{ik}) & \text{if } z_{jk} \geq z_{ik} \end{cases}$$

And $z_{ik} - z_{jk} = \chi_{jk}, z_{ik} - z_{jk} = \chi_{ik}$.

Remark 5.3: according to the Bonifacio Llamazares(2017), when the $f_1, f_2, g_1$ and $g_2$ are nondecreasing function, the value function $\varphi_k(A_i, A_j)$ fit the weight consistency and monotonicity.

**4.1 key algorithm**

the method of generalized BUI TODIM can be implemented as the following algorithm:

step 1: for each $i, j \in N$ and $k \in M$, we can calculate the dominance degree

$\varphi_k(A_i, A_j)$ by using the value function mentioned above

$$\varphi_k(A_i, A_j) = \begin{cases} m^{-1}\sum_{i=1}^{m}(\chi_{jk})^{\alpha}(w_k)^{\beta} & if\ p(z_{ik} \geq z_{jk}) > 0.5 \\ 0 & if\ p(z_{ik} \geq z_{jk}) = 0.5 \\ \theta^{-1}m^{-1}\sum_{i=1}^{m}(\chi_{jk})^{\alpha}(w_k)^{-\beta} & if\ p(z_{jk} \geq z_{ik}) > 0.5 \end{cases}$$

while the $\chi_{jk}$ inside is difference of input BUI value and $f_1, f_2, g_1, g_2$ inside are nondecreasing function.

In light of $w_k$ is a real number and the $\chi_{jk}, \chi_{ik}$ is BUI value, we note datum $x_i$ of $\chi_{jk}, \chi_{ik}$ as $x_{ik}$ or $x_{jk}$ and the part of a datum $c_i$ can be marked as $c_{ik}, c_{jk}$.

definition k: assume that aggregation function $A, A':[0,1]^n \to [0,1]$ is a weighted arithmetic mean associated with value function $f_1(x)$ and $f_2(x)$:

$$\begin{cases} A = \dfrac{1}{m}\sum_{k=1}^{m} f_1(x) \\ A' = \dfrac{1}{m}\sum_{k=1}^{m} f_2(x) \end{cases}$$

based on that, we define $A, A'$ to a more extensive form by establishing the association with relative criteria function $g_1(w_k)$ and $g_2(w_k)$, which is:

$$\begin{cases} A_w = \dfrac{1}{m}\sum_{k=1}^{m} g_1(w_k)f_1(x) \\ A_w' = \dfrac{1}{m}\sum_{k=1}^{m} g_2(w_k)f_2(x) \end{cases} \qquad A_w, A_w':[0,1]^n \to [0,1]$$

herein, we can define a mapping

$$\tilde{A}_L(<x_1;c_1>,\cdots,<x_1;c_1>) = <A(x_1,\cdots,x_n); 1-A(1-c_1+c_1x_1,\cdots,1-c_n+c_nx_n) + A(c_1x_1,\cdots,c_nx_n)>$$

and we can prove that $\tilde{A}_L$ it is an aggregation function related to $A$. (appendix A)

thus, we can define a novel aggregation function $\tilde{A}_{W,L}$ based on definition k,k definition k: let $A = A_w$ or $A = A'_w$ be a fixed aggregation function associated with relative criteria weight function $g_2(w_k)$ and $g_1(w_k)$. then, for any $x_i \in [0,1]^n$, we propose that

$$\tilde{A}_{w,l}(\chi_{jk}) =< A_w(x_{\chi_{jk}}); A_{w,x}(c_{\chi_{jk}}) >= m^{-1}\varphi(A_i, A_j)$$

herein, $A_{w,x}(c) = m^{-1}(1 - \sum_{k=1}^{m} g(w_k) \cdot A_w(1 - c_i + c_i x_i) + \sum_{k=1}^{m} g(w_k) \cdot A_w(c_i x_i))$

$\Rightarrow \varphi(A_i, A_j) = m\tilde{A}_{w,l}(\chi_{jk})$

And we can prove that the mapping $\tilde{A}_{w,l}(\chi_{jk}) = \tilde{A}_{w,l}(< x_1, c_1 >, \cdots, < x_m, c_m >)$ is an m dimensional BUI aggregation function related to nondecreasing function $f_1(x)$ and $f_2(x)$.

Proof: apparently, the mapping $\tilde{A}_{w,l}$ acting on the BUI sequences $\chi_{jk}$ is related to the aggregation function A (based on Definition k). Obviously $A = A_w$ or $A = A'_w$ is independent of $c_{\chi_{jk}}$ which implied definition of type-1 BUI aggregation satisfied. For any fixed $x_{\chi_{jk}} = (x_1, \cdots, x_n) \in [0,1]^n$, if $c_1 = c_2 = \cdots = c_n = 0$, then

$$1 - \sum_{k=1}^{m} g(w_k) \cdot A_w(1 - c_i + c_i x_i) + \sum_{k=1}^{m} g(w_k) \cdot A_w(c_i x_i)$$
$$= 1 - \sum_{k=1}^{m} g(w_k) \cdot A_w(0, \cdots, 0) + \sum_{k=1}^{m} g(w_k) \cdot A_w(1, \cdots, 1) = 0$$

therefore, $A_{w,x}(0, \cdots, 0) = 0$. Following the same principle,

$$A_{w,x}(1, \cdots, 1) = 1 - \sum_{k=1}^{m} g(w_k) \cdot A_w(1, \cdots, 1) + \sum_{k=1}^{m} g(w_k) \cdot A_w(1, \cdots, 1) = 1.$$

in light of the increasingness of $A_w$ in each coordinate, $A_{w,x}$ is also increasing in each coordinate, and thus is an aggregation function. In the sense of that, evidently $\tilde{A}_{w,l}$ satisfies properties of type-3 BUI aggregation function in Definition k. consequently, the mapping $\tilde{A}_{w,l}$ given by proposition k can be proved as a BUI aggregation function related to $A_w$ (as same as related to nondecreasing function $f_1(x)$ and $f_2(x)$).

step 3: calculating the overall performance of alternative $A_i$, $\varphi(A_i) = \sum\limits_{j=1}^{n} \varphi_k(A_i, A_j)$

let $\tilde{B}$ be an aggregation function related to the arithmetic mean $B = n^{-1} \cdot \sum\limits_{j=1}^{n} x_i$, which is $\tilde{B}(<x_1, c_1>, \cdots, <x_n, c_n>) = <B(p_x(\varphi(A_i, A_j))); B(p_c(\varphi(A_i, A_j)))>$

in the sense of definition k we can derive that

$$n^{-1} \cdot \varphi(A_i) = n^{-1} \sum\limits_{j=1}^{n} \varphi(A_i, A_j) = \tilde{B}(\varphi(A_i, A_j))$$

Herein, mapping $\tilde{B}$ can also be proven as a n-dimensional BUI aggregation function proof: if $c_1 = c_2 = \cdots = c_n = 0$, then $B<0,\cdots,0> = 0, B<1,\cdots,1> = 1$ can be easily found out. Moreover, the mapping $\tilde{B}(<x_1, c_1>, \cdots, <x_n, c_n>)$ acting on the BUI's is related to the arithmetic mean $B$ due to definition k, which also implies that BUI axiom 3 is satisfied. At last, it's clear to see that $B$ is independent of each $p_c(\varphi(A_i, A_j))$ implied BUI axiom one satisfied. Thus, the mapping $\tilde{B}$ given by proposition k can be proved as a BUI aggregation function related to $B$.

in a more generalized perspective, the equation k can be written as:

$$\varphi(A_i) = n \cdot \tilde{B}(m\tilde{A}_{w,l}(\chi_{jk})) = mn \cdot \tilde{B}(\tilde{A}_{w,l}(\chi_{jk}))$$

step 4: normalizing overall performance $\varphi(A_i)$

$$\xi(A_i) = \frac{\varphi(A_i) - \min_{j \in N} \varphi(A_j)}{\max_{j \in N} \varphi(A_j) - \min_{j \in N} \varphi(A_j)} = \frac{\tilde{B}(\tilde{A}_{w,l}(\chi_{jk})) - \min_{j \in N} \tilde{B}(\tilde{A}_{w,l}(\chi_{jk}))}{\max_{j \in N} \tilde{B}(\tilde{A}_{w,l}(\chi_{jk})) - \min_{j \in N} \tilde{B}(\tilde{A}_{w,l}(\chi_{jk}))}$$

step 4:

utilizing the possibility degree of BUI value to obtain the partial order of each normalized overall performance of alternative $A_i$ and $A_j$.

hence, the weak order of alternatives can be defined by

$$A_i \succ A_j \Leftrightarrow \xi(A_i) \succ \xi(A_j)$$

## 5. Case study

For more intuitively understanding and demonstrating the construes of BUI-GTODIM method, which is not only considering the hesitation of phycological behaviors of decision makers and also avoid two paradox happening, we will propose some cases based on recent outbreak pandemic caused by covid-19 in this section by utilizing the BUI-GTODIM method contrast to some related methods.

the outbreak of Covid-19 has threatened thousands of people in different countries and drastically slow down the development of local economy. in china, an emergency team composed by some experts had been initialized by the CDC department for coping with these challenging biosafety affairs. thus, accurately estimate the biosafety level of each is very crucial for virous subsequent decision making. for evaluating the most safety city among the central capital provinces, the emergency team with three DM quickly developed the following three crisis evaluation criterion index through an analysis of the situation and previous experience gained from the H1N1 virus avian flu outbreak. without losing any generality, we assuming each DM has the equivalence weight which is $\{\frac{1}{3}, \frac{1}{3}, \frac{1}{3}\}$

$A_1$: infectious disease monitoring equipment to the lockdown areas, moving the quarantined people to safer zones. Doing so allows the medical professionals to directly treat the infected and reduce the spread of disease at the scene of the containment, but the cost is relatively high.

$A_2$: Rescue buses are allowed to transport the relevant medical workers and small rescue equipment to the disaster site, moving people who are not yet affected to safe zones outside of the city. The cost of this alternative is relatively low.

$A_3$: Building special infectious disease hospitals to house the infected and contain the spread the disease. Doing so will take time and the cost of bringing the construction equipment and people is high, without a sense of how many people will be affected and the spread of the infection.

The following indicators need to be considered when DMs make decisions: prevention of secondary infectious disease spread($C_1$), the safety of medical personnel brought in to the lockdown area ($C_2$), the best evacuation process and route to take($C_3$), and the state of economic growth($C_4$).

Using equation mentioned above, the weighting vector of the attribute is $\omega=(0.2453,0.2571,0.2523,0.2453)^T$

To solve this problem in the emergency management caused by the COVID-19 incident, we apply the steps presented in the earlier sections of this paper. The following are the solution process and the results. the aggregated decision matrix of DMs can be obtained, as shown in Table 1.

Table1:Decision Making Matrix of DMs

|  | $C_1$ | $C_2$ | $C_3$ | $C_4$ |
|---|---|---|---|---|
| $A_1$ | <0.337;0.726> | <0.357;0.815> | <0.736;0.624> | <0.573;0.699> |
| $A_2$ | <0.336;0.721> | <0.275;0.682> | <0.719;0.720> | <0.546;0.733> |
| $A_3$ | <0.341;0.737> | <0.315;0.745> | <0.744;0.673> | <0.528;0.750> |

Using equation mentioned above, the weighting vector of the attribute is $\omega = (2453, 1571, 2454, 1678)^T$

Then, using mentioned above, the dominance degree matrix can be obtained as

$$\varphi(A_i) = \begin{bmatrix} <0.367; 0.768> & <0.297; 0.850> & <0.787; 0.718> & <0.537; 0.765> \\ <0.353; 0.710> & <0.394; 0.661> & <0.710; 0.868> & <0.573; 0.689> \\ <0.396; 0.754> & <0.410; 0.762> & <0.810; 0.668> & <0.691; 0.801> \end{bmatrix}$$

Here, is set to one $\theta, \alpha, \beta$.

Further, by using partition exchange sort algorithm, the perceived dominance value can be calculated. Therefore, the rank of the alternatives is A3 > A2 > A1. This choice is fully in line with the actual emergency situation estimated by the CDC among the central Province of China and proves the validity and feasibility of the method to some extent.

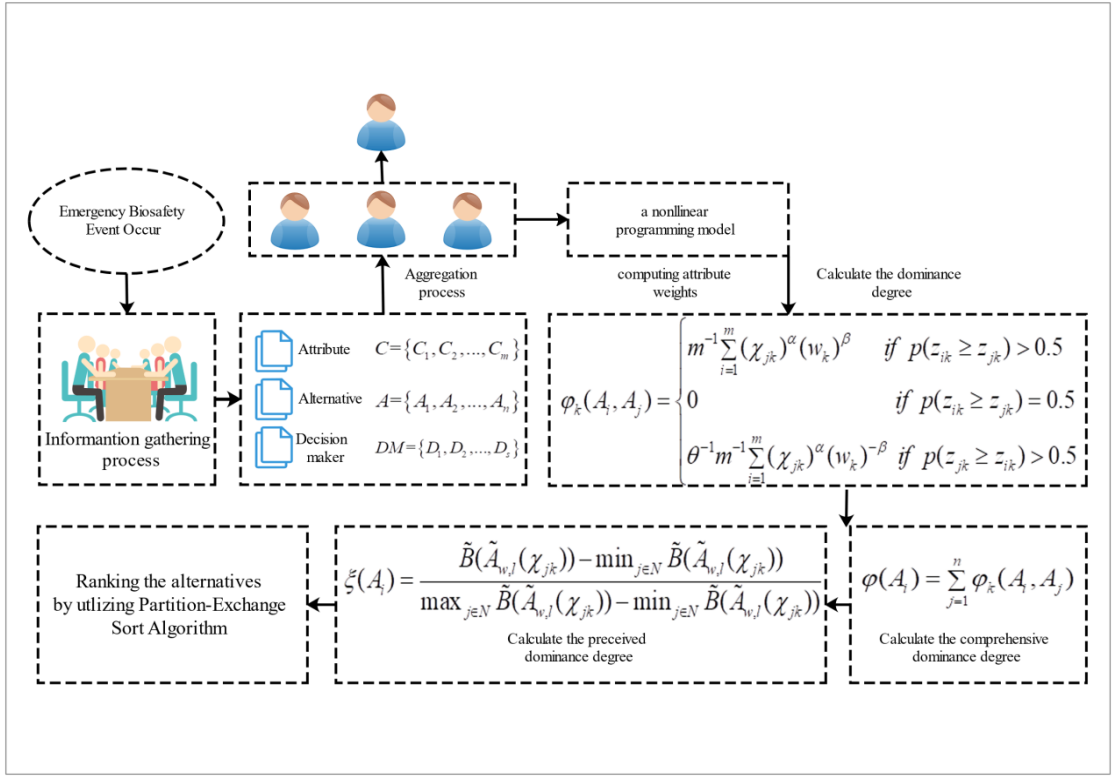

Fig. 2. Procedures for proposed method

## 6. Conclusions

This paper presents an extended TODIM approach for group emergency decision making based on bidirectional projection. This method takes into account the psychological behavior of the DMs during the decision-making process, which is currently under studied. We combine the TODIM method with a basic uncertain information to overcome the limitation of the traditional projection method, which cannot accurately rank the solution alternatives. Therefore, the proposed method is quite different from the existing GEDM method and provides a better decision making outcome. In the proposed method, because of the incomplete information and complexity of the emergencies, the DMs use the BUI to express their personal preference information, which is better than crisp numbers. The weight of attributes is used to make a fair comparison with the existing related method in the case study to demonstrate the novelty, feasibility, and effectiveness of the proposed method. In future, to accommodate complex decision environments, the individual preference value of DMs may be expressed by a variety of different fuzzy information, based on an uncertain environment.

**7.Reference**